\documentclass[sigconf]{acmart}

\usepackage{soul}
\usepackage[utf8]{inputenc}
\usepackage{graphicx}
\usepackage{amsmath}
\usepackage{amsthm}
\usepackage{booktabs}
\usepackage[inline]{enumitem}
\usepackage{multirow}
\usepackage[capitalize]{cleveref} 
\usepackage{listings}
\usepackage{color}
\usepackage{xspace} 
\usepackage{subfig} 
\newcommand{\PBAMnet}{{\sc{pFoodReQ}}\xspace}
\newcommand{\BlackPBAMnet}{{\sc{\textbf{pFoodReQ}}}\xspace}
\newcommand{\PBAMnetSim}{{\sc{pFoodReQ+RecipeSim}}\xspace}

\usepackage{float}
\usepackage{mathrsfs} 
\let\vec\mathbf

\usepackage{xparse}
\usepackage{xcolor}
\NewDocumentCommand{\zaki}{ mO{} }{\textcolor{red}{\textsuperscript{\textit{Zaki}}\textsf{\textbf{\small[#1]}}}}
\NewDocumentCommand{\chinghua}{ mO{} }{\textcolor{green}{\textsuperscript{\textit{Ching-Hua}}\textsf{\textbf{\small[#1]}}}}
\NewDocumentCommand{\ananya}{ mO{} }{\textcolor{cyan}{\textsuperscript{\textit{Ananya}}\textsf{\textbf{\small[#1]}}}}
\NewDocumentCommand{\hugo}{ mO{} }{\textcolor{blue}{\textsuperscript{\textit{Hugo}}\textsf{\textbf{\small[#1]}}}}
\AtBeginDocument{%
  \providecommand\BibTeX{{%
    \normalfont B\kern-0.5em{\scshape i\kern-0.25em b}\kern-0.8em\TeX}}}

\copyrightyear{2021}
\acmYear{2021}
\setcopyright{acmcopyright}\acmConference[WSDM '21]{Proceedings of the
   Fourteenth ACM International Conference on Web Search and Data
   Mining}{March 8--12, 2021}{Virtual Event, Israel}
\acmBooktitle{Proceedings of the Fourteenth ACM International Conference
   on Web Search and Data Mining (WSDM '21), March 8--12, 2021, Virtual
   Event, Israel}
\acmPrice{15.00}
\acmDOI{10.1145/3437963.3441816}
\acmISBN{978-1-4503-8297-7/21/03}

\begin{document}

\title{Personalized Food Recommendation as Constrained Question Answering over a Large-scale Food Knowledge Graph}


\author{Yu Chen$^1$, Ananya Subburathinam$^1$, Ching-Hua Chen$^2$, Mohammed J. Zaki$^1$}
\email{hugochan2013@gmail.com, subbua@rpi.edu, chinghua@us.ibm.com, zaki@cs.rpi.edu}
\affiliation{$^1$ Rensselaer Polytechnic Institute, Troy, NY
$\qquad$
$^2$IBM Research, Yorktown Heights, NY
}






\begin{abstract}
Food recommendation has become an important means to help guide users to adopt healthy dietary habits. Previous works on food recommendation either i) fail to consider users' explicit requirements, ii) ignore crucial health factors (e.g., allergies and nutrition needs), or iii) do not utilize the rich food knowledge for recommending healthy recipes. To address these limitations, we propose a novel problem formulation for food recommendation, modeling this task as constrained question answering over a large-scale food knowledge base/graph (KBQA). Besides the requirements from the user query, personalized requirements from the user's dietary preferences and health guidelines are handled in a unified way as additional constraints to the QA system. To validate this idea, we create a QA style dataset for personalized food recommendation based on a large-scale food knowledge graph and health guidelines. Furthermore, we propose a KBQA-based personalized food recommendation framework which is equipped with novel techniques for handling negations and numerical comparisons in the queries. Experimental results on the benchmark show that our approach significantly outperforms non-personalized counterparts (average 59.7\% absolute improvement across various evaluation metrics), and is able to recommend more relevant and healthier recipes.
\end{abstract}



\maketitle

\section{Introduction}

Many chronic diseases (e.g., type 2 diabetes) are related to poor eating habits~\cite{DeSalvo2016}.
Diet management is hence becoming indispensable in our daily lives.
However, how to translate personalized health conditions and health knowledge into day-to-day dietary practices remains a challenging problem.
Food recommendation has become an important means to help guide users to adopt healthy dietary habits~\cite{ge2015health, forbes2011content,Teng2012,ueda2014recipe,Trattner2017,yang2017yum,min2019food}.
Most existing methods recommend recipes based on recipe content (e.g., ingredients) or user behavior history (e.g., eating history).
Following the classical paradigm of recommendation systems, they typically treat the food recommendation problem as a matrix completion or classification problem.

However, previous methods have three main drawbacks.
First, they mostly do not consider users' explicit requirements (e.g., ``breakfast dishes with bread'') when doing food recommendation.
Second, they mainly focus on recommending recipes which comply well with users' eating habits without considering crucial health factors such as allergies and nutrition needs.
This might have a potential risk of encouraging users to eat similar but unhealthy recipes, especially if users already have poor eating habits.
Third, most of them fail to utilize the rich food knowledge for recommending healthy recipes. There are many available recipe datasets providing ingredient lists and other recipe information that can be utilized for the purpose of food recommendation, such as ~\cite{salvador2017learning}, Yummly (\url{www.yummly.com}), and Allrecipes (\url{www.allrecipes.com}).
Recently, Haussmann et al.~\shortcite{haussmann2019foodkg} released a large-scale and unified food knowledge graph (KG), namely FoodKG, which brings together food ontologies, recipes, ingredients and nutritional data. Resources like this can be used for more effective food recommendation.

In this work, we propose to treat the task of personalized food recommendation as constrained question answering over a food KG, which manages to address the above limitations in a unified way.
Specifically,
recommendation is conducted in a Knowledge Base Question Answering (KBQA)~\cite{bordes2014open} setting
where a KBQA system takes as input a user query (e.g., ``what is a good breakfast that contains bread?'') and retrieves all recipes from the KG 
that satisfy the query.
This allows user queries to be posed more naturally.
Our approach further offers recommendations that comply well with  ``personalized'' requirements based on a user's persona, which includes the user's unique health conditions (e.g., allergies),  applicable health guidelines (nutrition needs), and even food logs.
We propose a personalized KBQA approach, \PBAMnet for {\em {\bf P}ersonalized {\bf Food} {\bf Re}commendation via {\bf Q}uestion answering} (pronounced as `FoodRec'), which regards such personalized requirements as additional constraints. 
In order to adapt regular KBQA approaches~\cite{chen2019bidirectional} to handle the personalized KBQA setting,
novel techniques are proposed to 
append personalized requirements to the raw user query via \emph{query expansion},
handle numerical comparisons via \emph{KG augmentation},
and negations via \emph{constraint modeling}. 
Lastly, to validate our approach, we create 
a QA style food recommendation benchmark based on  FoodKG~\cite{haussmann2019foodkg} and American Diabetes Association (ADA) lifestyle guidelines~\cite{american20195}.
Our main contributions are as follows:
\vspace{-\topsep}
\begin{itemize}
  \setlength{\parskip}{0pt}
  \setlength{\itemsep}{0pt plus 1pt}
  
\item We propose a KBQA-based personalized food recommendation framework 
 that regards personalized requirements from dietary preferences and health guidelines as additional constraints to QA.
 To the best of our knowledge, we are the first to propose the QA-based strategy for food recommendation.

\item Novel techniques including \emph{KG augmentation} and \emph{Constraint modeling} are proposed to effectively handle numerical comparisons and negations in the queries. 

\item We create a QA style benchmark for personalized food recommendation based on a large-scale food knowledge graph and health guidelines. 

\item Experimental results show that our approach significantly outperform non-personalized counterparts (average 59.7\% absolute improvement across various evaluation metrics) and is able to recommend more relevant and healthier recipes.

\end{itemize}

\section{Related Work}

Recommendations systems abstract away the myriad of choices and decisions that have to be made before arriving at the choice that caters to the needs and preferences of an individual. To this end, they have been deployed successfully in many diverse domains from recommending movies ~\cite{li2019haes, diao2014jointly} and music ~\cite{jin2019contextplay, cheng2017exploiting} to e-commerce ~\cite{wang2018billion}
and healthcare ~\cite{wiesner2010adapting, hu2016personal}.

\subsection{Food Recommendation}
Most existing food recommendation approaches recommend recipes based on recipe content (e.g., ingredients)~\cite{forbes2011content,Teng2012,el2012food,chen2020eating}, user behavior history (e.g., eating history)~\cite{forbes2011content,ueda2014recipe} or dietary preferences~\cite{ueda2014recipe}. Based on their survey, \cite{ge2015health} established that most users would be willing to adapt their taste in favour of healthier recipe options, and developed a health-aware recommender algorithm that takes personal profile information and stated preferences into account when making recipe recommendations. 
Additionally, recipe recommendation websites and apps also use some combination of user-selected ingredients, user history, or certain food tags (such as ``Italian'', or ``kid-friendly'') to retrieve and recommend recipes. ChooseMyPlate.gov (\url{https://www.choosemyplate.gov/myplatekitchen/recipes}) includes nutrition focus criteria, cost, and cooking equipment as additional filters to their recipes. Unlike these services, {\em our work improves personalization by automatically applying persona focused nutritional guidelines and ingredient constraints to every query, in addition to allowing recipe queries to be posed as natural language questions}.

Following the classical paradigm of recommendation systems,
most existing methods treat the food recommendation problem as a matrix completion or classification problem.
To evaluate a recommendation system, they take recipe ratings, or whether a user selects the recommended recipe or not, as ground-truth.
Our method is more user-friendly and has the potential to be used for interactive or conversational food recommendation, for example, in a personalized health recommendation app. 
Furthermore, various requirements (e.g., dietary preferences) can be addressed in a unified way as constraints to a QA system.
The most similar work to ours is~\cite{haussmann2019foodkgdemo}. 
Whereas they presented a demo QA application that answers fact based simple, comparison and constraint-based questions, their work does not tackle personalization. In contrast, we explicitly focus on personalized food recommendation.

\subsection{Knowledge Base Question Answering}

The task of knowledge base question answering (KBQA) is to automatically find answers from an underlying KG given a natural language question.
KBQA approaches fall into two broad  categories: semantic parsing (SP) based \cite{wong2007learning,krishnamurthy2012weakly,cai2013large,yih2015semantic,abujabal2017automated,hu2018answering} or information retrieval (IR) based 
\cite{bordes2014open,bordes2014question,dong2015question,bordes2015large,xu2016question,hao2017end,chen2019bidirectional}.
Semantic approaches typically transform questions into intermediate logic forms, which can be posed as queries to a KG.
However, they require a pre-defined set of lexical triggers or rules, which limits their domain of applicability and also their scalability.
Retrieval based methods directly retrieve answers from a KG based on the information from the questions. 
They usually do not require hand-made rules and can therefore scale better to large and complex knowledge graphs.
These methods span many embedding-based methods, that focus on mapping answers and questions into a common embedding space, where one can query the KG independent of its schema, without requiring any grammar or lexicon. 
We extend IR based KBQA methods to the personalized setting, where personal and contextual information must be utilized so as to find correct answers from a KG.

\section{Approach}

\subsection{Personalized KBQA Benchmark}\label{sec:benchmark_creation}

We create a benchmark QA dataset based on the extensive FoodKG \cite{haussmann2019foodkg} knowledge graph that contains over 1 million recipes, along with their ingredients and nutrients, and using ADA lifestyle guidelines~\cite{american20195}. 
We refer the interested readers to~\cite{haussmann2019foodkg} for details about the FoodKG.
To the best of our knowledge, this is the first dataset that supports QA style personalized food recommendations related to ingredients, nutrients and recipes.
Each example in the dataset contains a \emph{user query}, \emph{dietary preferences}, \emph{health guidelines} associated with the user, and the \emph{ground-truth answers} (i.e., recipe recommendations).
Note that the ground-truth answers in our benchmark are those recipes from FoodKG that satisfy both explicit requirements (mentioned in the user query) and personalized requirements (mentioned in the dietary preferences or health guidelines).
The details on how we create the benchmark are given next.

\begin{figure*}[!ht]
    \centering
    \vspace{-0.2in}
    \centerline{
    \subfloat[Simple Query]{
    \includegraphics[
    width=4.5in
    ]{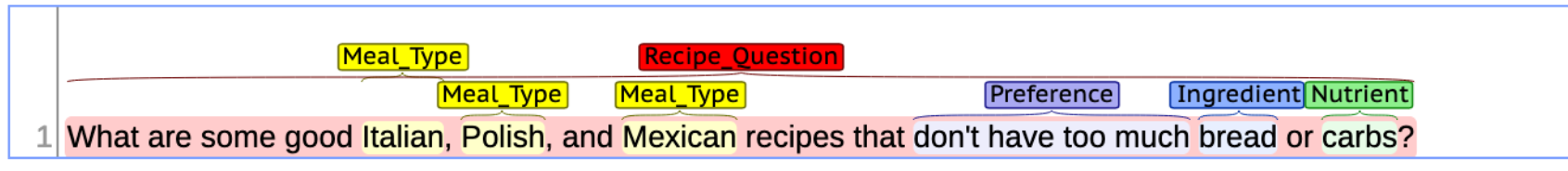}
    }}
    \vspace{-0.1in}
    \centerline{
    \subfloat[Ingredient Preference Query]{
    \includegraphics[
    width=5.5in, height=1.75in
    ]{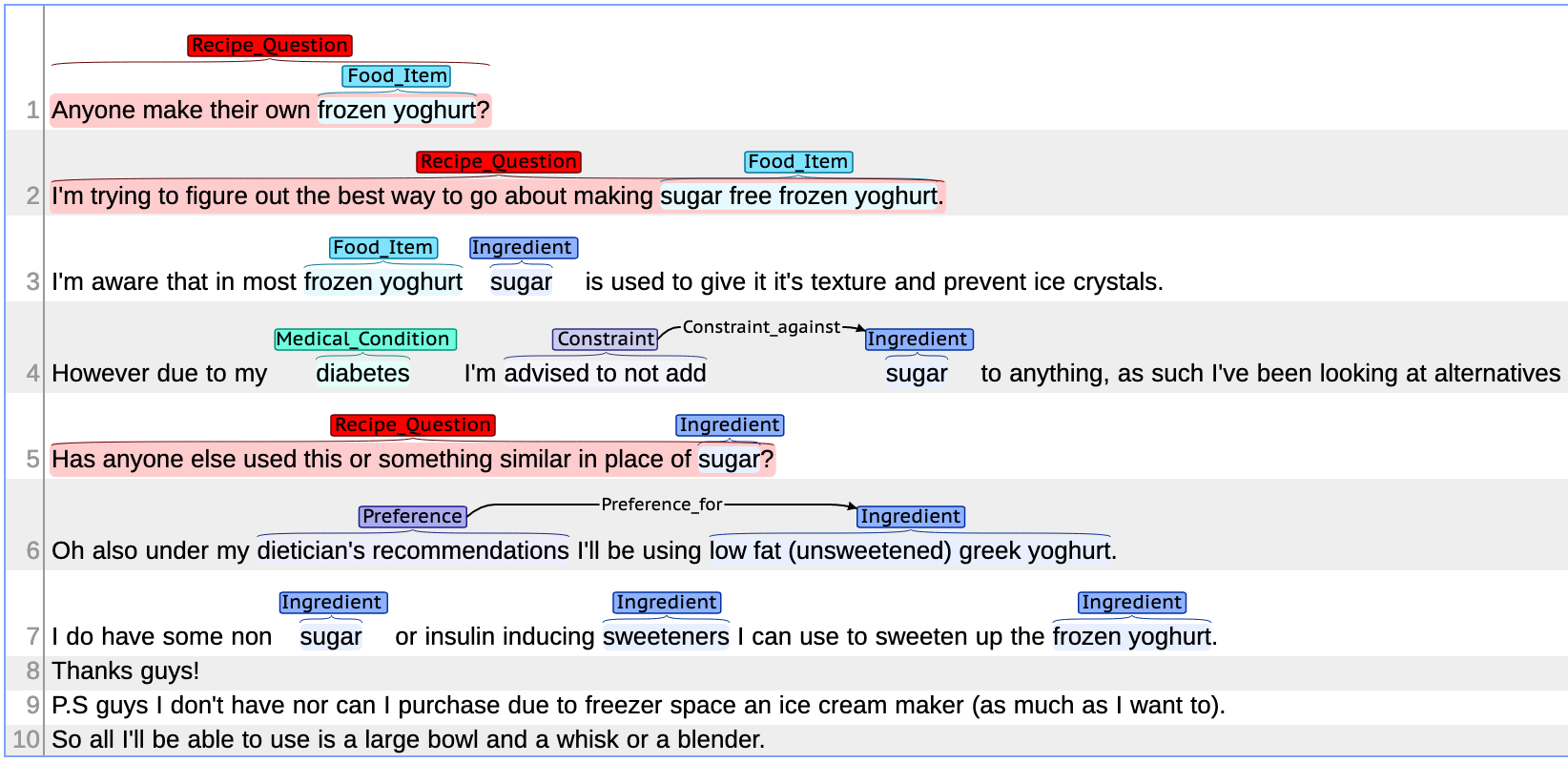}
    }}
    \vspace{-0.1in}
    \centerline{
    \subfloat[Ingredient Exclusion Query]{
    \includegraphics[
    width=6in, height=1.5in
    ]{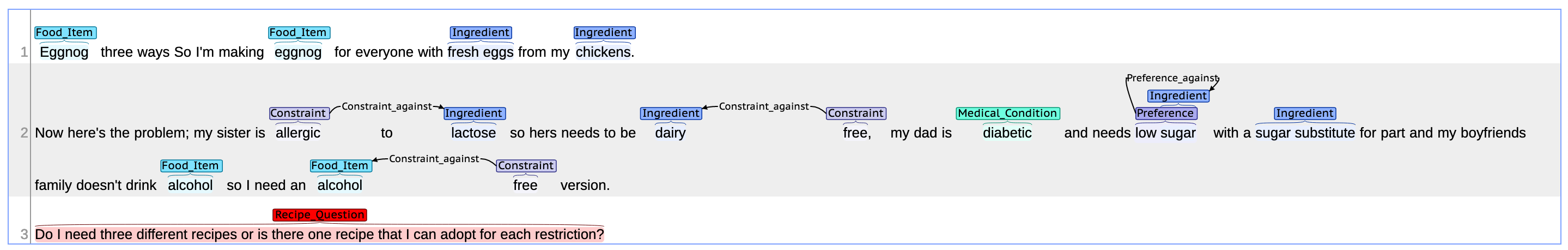}
    }}
    \vspace{-0.1in}
    \caption{Examples of real user questions on Reddit annotated with food items, ingredients, nutrients, allergy/preference constraints and other tags. (a) Simple query about cuisine type and ingredient/nurtient constraints. (b) Query illustrating various preferences and dietician-guided recommendation. (c) Query indicating various constraints related to allergy, medical condition and diet restriction.}
    \label{fig:reddit_examples}
    \vspace{-0.1in}
\end{figure*}


\subsubsection{Template-based Natural Language Question Generation} 

We create natural language questions that ask for food recommendation with explicit requirements.
To generate realistic benchmark questions that
reflect real word requirements and constraints, we examined over $200$ submissions related to recipe and diabetes made on the sub-reddits ``AskCulinary'', ``Cooking'', ``1200isplenty'', ``Baking'', ``AskDocs'', ``AskScience'', and ``AskReddit'' on the social media forum, Reddit (\url{http://www.reddit.com/}). 
Figure~\ref{fig:reddit_examples} shows some examples of real user queries on reddit, which we have annotated for mentions of ingredients, nutrients, food items, as well as stated preferences and constraints for or against ingredients or nutrients.
From the reddits, we identified a total of $156$ posts that ask for recipe suggestions. Out of these, Table~\ref{table:reddit_stats} shows the breakdown of how many posts mention specific ingredients, specific nutrients (‘fat’ , ‘carbs.’), food items, meal type, or preferences/constraints for/against specific ingredients/nutrients.
Inspired by these real user queries, in our benchmark we generate questions that include the following four common types of constraints found in the reddit submissions:
i) positive ingredient constraints stating what ingredient(s) can be included in the recipe, 
ii) negated ingredient constraints stating what ingredient(s) cannot be included, 
iii) nutrient based constraints such as ``low carb'' or ``high protein'', and 
iv) cuisine based constraints such as ``Indian'', or  ``Mediterranean''. 

\begin{table}[!ht]
\vspace{-0.15in}
\begin{center}
\caption{Statistics from 156 recipe requests from Reddit.}
\label{table:reddit_stats}
\vspace{-0.1in}
\scalebox{1}{
\begin{tabular}{lll}
\toprule
Mention Type &  Number of Posts \\
  \midrule
Ingredient & 112\\
Nutrient & 48\\
Food Item & 109\\
Meal Type & 51\\
Preference/Constraint & 104\\
\bottomrule
\end{tabular}
}
\end{center}
\vspace{-0.1in}
\end{table}

Nutrient based limits in the templates are randomly generated from a combination of [low/medium/high] [fat/protein/carbohydrates]. The thresholds defining low, medium, and high are set for each of these nutrients based on established online sources~\cite{healthcom, clevelandclinic, zelman_2018}. We also accommodate queries that feature a combination of these constraint types. Example: "What \{limit\} \{nutrient\} \{tag\} dishes can I make that do not contain \{in\_list\}?" We have 56 different template questions in total that we use to generate the benchmark questions, that span the main types of questions gleaned from the reddit submissions.
Table~\ref{table:question_templates} shows some example templates, and the questions generated from them.

\begin{table*}[!htb]
\vspace{-0.1in}
\small
\centering
\caption{Examples of questions templates and generated questions (guided by real user questions on social media).} 
\label{table:question_templates}
\vspace{-0.1in}
\begin{tabular}{{p{9cm} @{\hspace*{2mm}}p{8cm}}}
\toprule
  {\bf Template} & {\bf Generated Question}\\
  \midrule
  What are \{tag\} recipes that contain \{in\_list\}? &  What are jellies recipes that contain orange?\\ 
  
  What \{tag\} recipes can I cook without \{in\_list\}? & What turkish or dinner-party recipes can I cook without canned milk?\\  
  
  Recommend \{limit\} \{nutrient\} \{tag\} recipes which have  \{in\_list\}? & Recommend low protein russian recipes which have  onions?\\  
  
  Suggest \{limit\} \{nutrient\} \{tag\} dishes which don't contain \{in\_list\}? & Suggest low fat egyptian dishes which don't contain lean ground beef?\\ 
  \bottomrule
\end{tabular}
\vspace{0.1in}

\end{table*}
\begin{table*}[!htb]
\vspace{-0.1in}
\centering
\caption{Data statistics of the personalized KBQA benchmark. The test set includes in-domain and out-of-domain instances.}
\label{table:data_statistics}
\vspace{-0.1in}
\begin{tabular}{l ll ll l}
\toprule
   & Train & Dev & Test (in-domain) & Test (out-of-domain) & Test\\
\midrule
 Size &  4,621 & 1,540  & 1542 & 727 & 2,269\\ 
 Avg. raw query len. & 11.6 &11.6 & 11.5 & 11.5 & 11.5  \\
 Avg. expanded query len. & 35.3 &35.1 & 35.1 & 35.1 & 35.1\\ 
 Avg. \# of answers &  3.2 & 2.9 & 2.6 & 3.0 & 2.7\\ 
 Avg. \# of constraints& 5.0 & 5.0 & 5.0 & 5.0 & 5.0\\ 
\bottomrule
\end{tabular}
\vspace{-0.1in}
\end{table*}

\begin{figure}[!ht]
    \centering
    \includegraphics[
    height=1.3in,
    width=2.5in
    ]{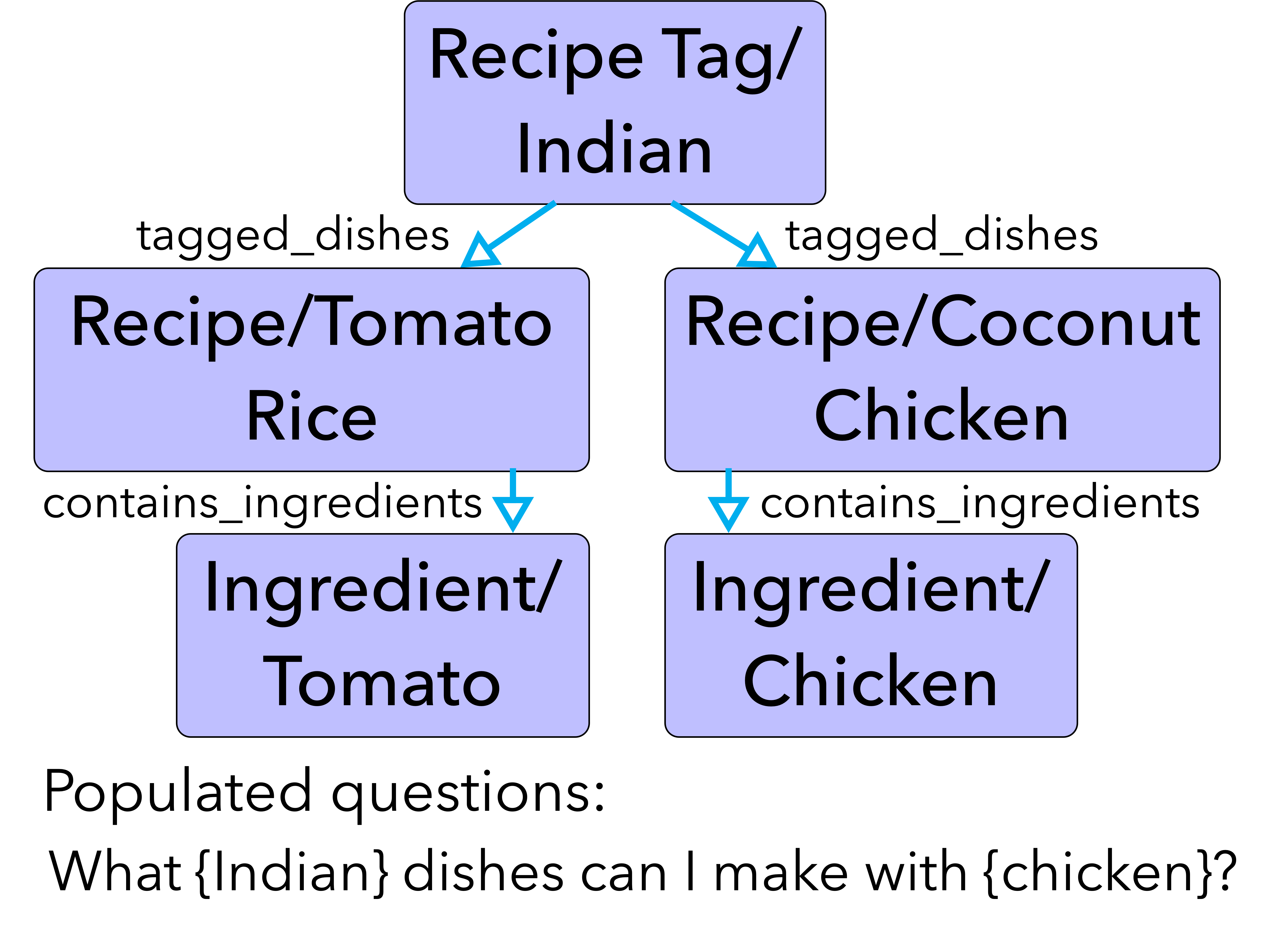}
    \vspace{-0.1in}
    \caption{Template-based natural question generation.}
    \label{fig:data_creation}
\end{figure}
\vspace{-0.1in}
    
As shown in Figure~\ref{fig:data_creation},
in order to generate a question such as ``what Indian dishes can I make with chicken?'',
we first randomly sample a subgraph from FoodKG which is defined as a $h$-hop neighborhood surrounding a food tag entity (e.g., ``Indian'').
We then randomly sample a question template (e.g., ``what [tag] dishes can I make with [ingredients]?'')
from the template pool,
and fill the slots with KG entities (e.g., ``chicken'') sampled from the subgraph. 

\subsubsection{Generating Dietary Preferences}
A good food recommendation system should respect a user's personalized requirements even though the user does not explicitly mention them in a query.
Keeping this in mind, to make our benchmark more realistic, 
associated to each user query,
we randomly generate dietary preferences on ingredient likes and allergies, where ingredients are randomly sampled from an ingredient pool that includes all the ingredients contained in the recipes belonging to the food tag mentioned in the query.
The average/min/max sizes of ingredient pools are 717, 58 and 2,093, respectively.
\begin{figure}[!ht]
  \vspace{-0.1in}
  \centering
  \includegraphics[
    keepaspectratio=true,
    scale=0.2,
  ]{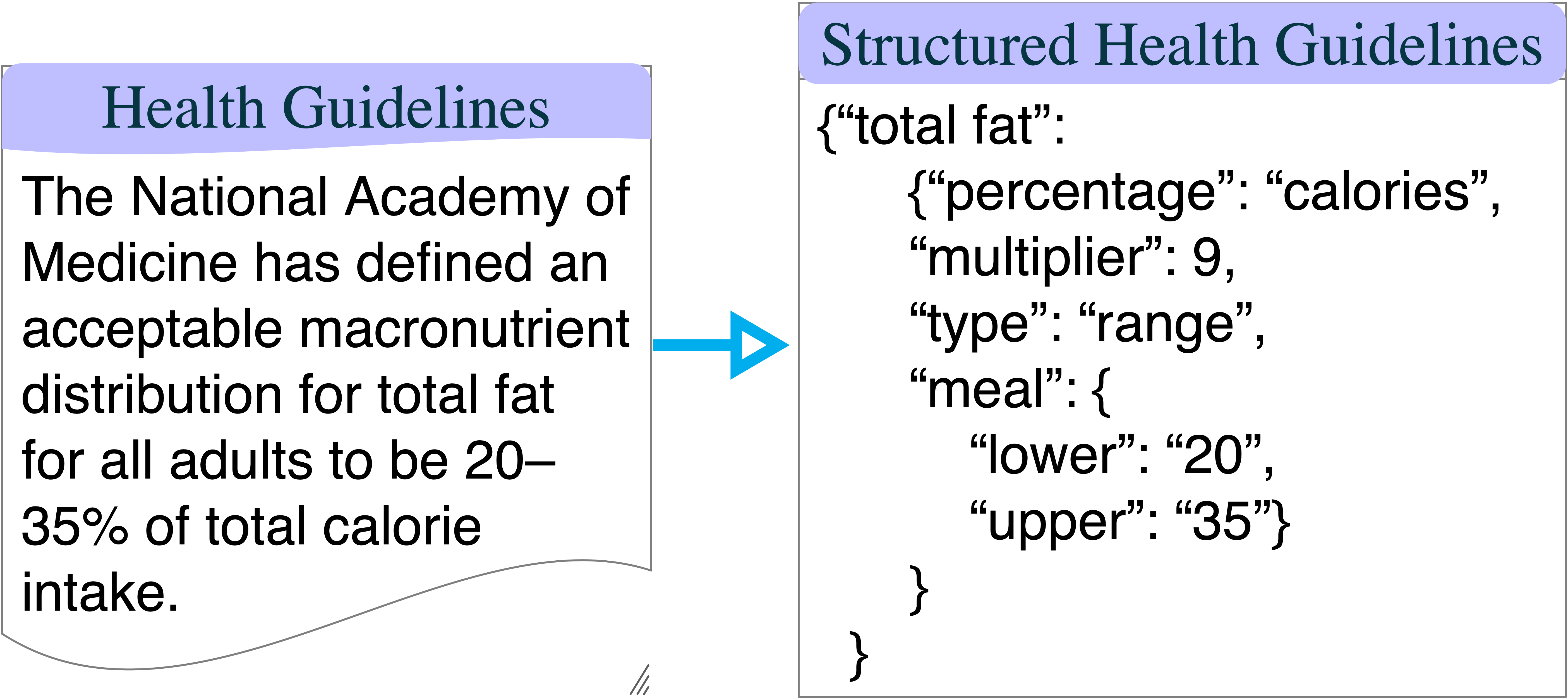}
    \vspace{-0.1in}
  \caption{A guideline example from the ADA.}
  \label{fig:guideline}
  \vspace{-0.2in}
\end{figure}

\subsubsection{Compiling Health Guidelines}

A characteristic that distinguishes our proposed method from most existing food recommendation methods is that
we want our system to recommend ``healthy'' recipes that comply with health guidelines.
Therefore, we carefully select some food-related guidelines from ADA lifestyle guidelines~\cite{american20195},
that pertain to nutrient and micronutrient budgets,
and treat them as additional requirements for food recommendation.
Since those guidelines are in natural language, we convert them into structured representations (e.g., a hash table storing key-value pairs).
For instance, as shown in Figure~\ref{fig:guideline},
the guideline suggests that a good percentage of calories from total fat should be between 20\% and 35\%.
This kind of macronutrient is calculated by dividing
the amount of calories from total fat by the amount of total calories in a recipe.
Note that the multiplier value in the example (right hand panel) indicates that ``food nutrient as fat (lipids) contains 9 kilocalories per gram (kcal/g)''.
Each dietary preference in the dataset is randomly assigned actual health guidelines from the ADA.
For more realistic setting, we allow multiple  guidelines per user; we set the number of guidelines per user to be 1, 2 or 3 with probabilities 0.85, 0.1 and 0.05, respectively, to avoid conflicting or over-restricting guidelines when creating the dataset.
As such, our framework is generic enough to handle several health guidelines. 

\subsubsection{KBQA Benchmark Dataset}
We split the personalized KBQA benchmark dataset into training (4,621 examples),
development (1,540 examples) and test sets (2,269 examples)
where 727 examples in the test set are out-of-domain examples.
Out-of-domain examples were created based on 23 recipe tags (out of totally 238 recipe tags) which were never seen in the training and development sets.
The data statistics are provided in Table~\ref{table:data_statistics}.

\subsection{\BlackPBAMnet: Personalized Food Recommendation Framework}

\begin{figure*}[!ht]
    \vspace{-0.1in}
    \centering
    \includegraphics[
    height=1.7in,
    width=5.6in]{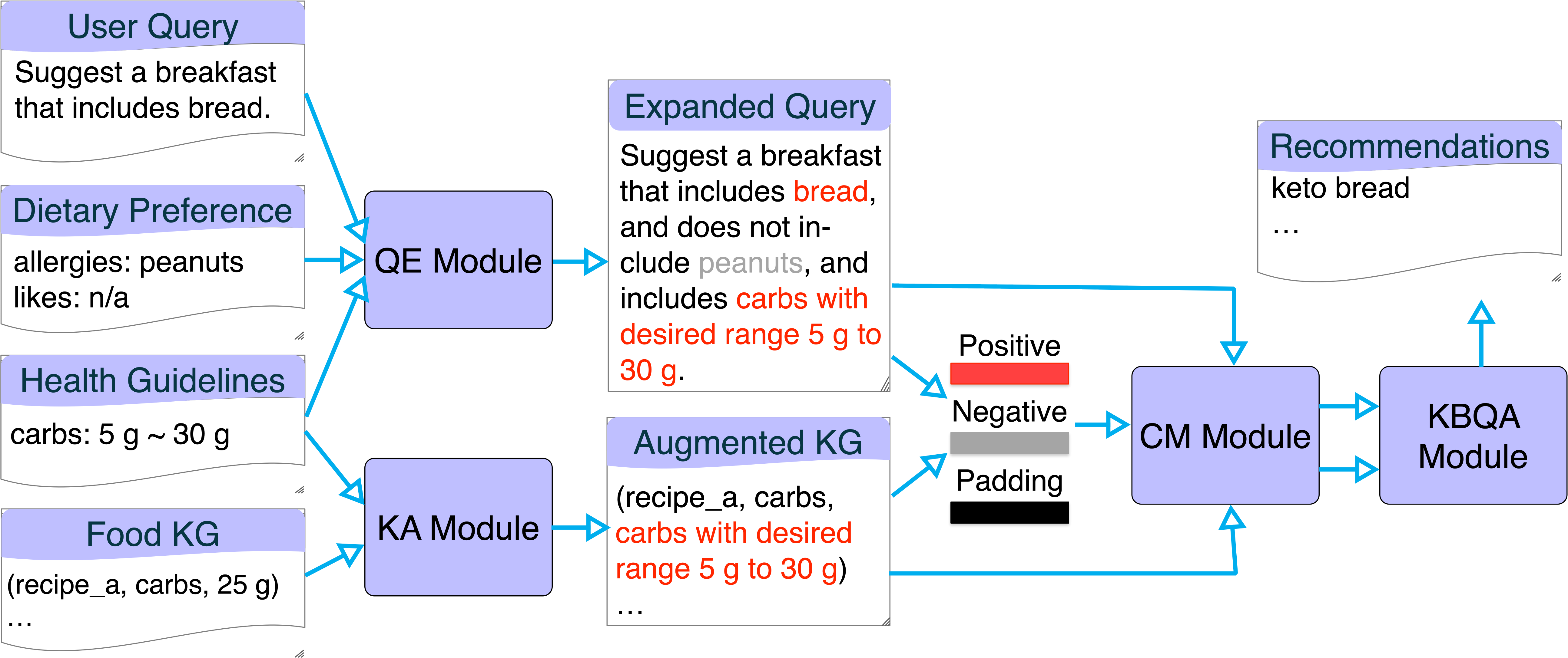}
\vspace{-0.1in}
    \caption{\PBAMnet architecture:
    QE, KA and CM stand for Query Expansion, KG Augmentation and Constraint Modeling, respectively. Positive constraints are in red, negative constraints in gray, and non-constraints (or padding) in black.}
    \label{fig:arch}
\vspace{-0.1in}
\end{figure*}

We treat the task of personalized food recommendation as constrained question answering over a food KG with additional personal constraints from dietary preferences and health guidelines.
Formally, we formulate the following problem setting:
a personalized food recommendation system is supposed to take as input a natural language question, dietary preferences  as well as health guidelines, and retrieve all recipes from a food KG that satisfy the requirements contained in the input.
The overall architecture of our proposed \PBAMnet framework is shown in Figure~\ref{fig:arch}, which consists of the \emph{Query Expansion (QE)}, \emph{KG Augmentation (KA)}, \emph{Constraint Modeling (CM)}, and \emph{KBQA} modules.

\subsubsection{KBQA Module}
The KBQA module is responsible for retrieving recipes from the KG that satisfy the constraints mentioned in the expanded query (will be discussed later).
Our approach to personalization is independent of the underlying KBQA method. 
We can use any KBQA method that accepts a query and a KG subgraph, and returns the most relevant answers from the KG subgraph.
As in most KBQA methods, a KG subgraph is extracted by first detecting a topic entity (i.e., the main entity in a query), and then fetching a $h$-hop neighborhood (which contains entities and relations) surrounding the topic entity. 
We will test our framework with various KBQA models in our experiments.


\subsubsection{Query Expansion}

An effective food recommendation system should respect personalized requirements from the dietary preferences and health guidelines.
As shown in Figure~\ref{fig:arch}, considering a user query ``suggest a breakfast that contains bread'', along with the raw user query, the dietary preferences indicate that the user is allergic to ``peanuts'', and the associated health guidelines suggest that the user consumes ``carbohydrates with desired range 5g to 30g''.
In order to make our system aware of the personalized requirements, we transform the raw user query into one that additionally incorporates the personalized requirements. 
Formally, given a number of personal constraints, say $\{c_1, c_2, ..., c_m\}$, from dietary preferences or applicable health guidelines, and the raw user query $q$ in natural language,
we append each of the constraints $c_i$ to $q$, and finally obtain the expanded query $\Tilde{q}$.
In the above example, the expanded query becomes ``suggest a good breakfast that contains bread, and does not have peanuts, and contains carbohydrates with desired range 5g to 30g''.
Thus, given the same query $q$,
our system is able to provide different users with tailored food suggestions that suit their needs based on the expanded query $\Tilde{q}$.

\subsubsection{KG Augmentation}

In order to answer a query with a numerical constraint like ``contains carbohydrates with desired range 5g to 30g'' from health guidelines, 
a KBQA system needs to handle numerical comparison. For example, in Figure~\ref{fig:arch}, the example \emph{recipe\_a} has $25g$ of carbohydrates, and therefore to check whether the recipe meets the guideline, the system would have to determine that``25 is larger than 5'' and ``25 is less than 30''.
Unfortunately, the inability to handle numerical comparison is a known limitation of neural networks.
Even though one can embed a number into a continuous vector,
regardless of the out-of-vocabulary issue,
it is extremely difficult to train a neural network to do reliable numerical comparisons in some high-dimensional embedding space.

We propose an effective technique that dynamically augments a KG subgraph, i.e., we create or {\em materialize} a new view of the KG subgraph, based on the results of symbolic number comparison before feeding it to a neural network-based KBQA system.
Let $G$ be the set of numerical constraints expressed in natural language (e.g., $g$ = ``carbohydrates with desired range 5g to 30g'', where $g \in G$). 
Let $v$ be the node in the KG subgraph that denotes the nutrition amount (e.g., the number of grams of carbohydrates) of a recipe. Then $v$ is replaced with $g$ if $v$ is within the desired numerical range (e.g., $[5, 30]$).
More formally, for each constraint $g \in G$, we model it as an indicator function $I_g(v) = 1$ if $v \in \text{range}(g)$; otherwise, $I_g(v) = 0$.
If $I_g(v)=1$, then $v$ is replaced with $g$; otherwise, we set it to empty.
In this way, no changes need to be made to the KBQA system to handle numerical comparisons.
Of course, when $I_g(v) = 0$, we can even completely remove the entire recipe node from the candidate set since it does not satisfy the constraint.
However, we choose to keep the recipe node because i) we want to train our KBQA system to retrieve the right answers on its own, which we think is more generic and ii) this opens a door for modeling soft constraints, which can can enhance the robustness and usefulness of our system, allowing it not to necessitate adherence to all of the constraints.  

\subsubsection{Constraint Modeling}

When dealing with queries involving ingredient allergies, a KBQA system should be able to handle negative constraints. 
However, it is well-known that neural networks are not good at handling negation~\cite{blanco2011semantic,fancellu2016neural,gautam2018long}.
A neural network-based KBQA system can have trouble distinguishing a negative constraint like ``does not have peanuts'' from a positive constraint like ``has peanuts''.
We propose an embedding-based method for constraint modeling.
The idea is to explicitly indicate the existence of a negative constraint to a KBQA system so that it can learn how to appropriately process the negative constraint through training.
Formally, when encoding an expanded user query $\Tilde{q}$, 
we add an additional constraint type markup $m_i \in M$ to each query word $q_i$ to indicate which type of constraint it belongs to.
$M$ is the set of constraint types that include {\em positive, negative and padding} where {\em padding} means the word does not belong to any constraint.
As shown in the expanded query of Figure~\ref{fig:arch}, words in black belong to the {\em padding} type as they are not part of the constraint set.
When encoding a candidate answer node from a KG subgraph,
we also add such constraint type markups to words occurring in its context node $v$ in order to explicitly indicate the constraint type to a KBQA system.
Note that all constraint type markups are linearly mapped into a continuous embedding space via a $|M| \times d_m$ weight matrix where $|M|$ is the size of the constraint set $M$ and $d_m$ is the dimension of the constraint embeddings.
The resultant constraint embeddings are then concatenated with the word embeddings before being further processed by a KBQA system.

\subsection{Training the Personalized KBQA system}\label{sec:kbqa_training}

As outlined in Figure~\ref{fig:arch},
our personalized KBQA system takes as input an expanded natural language question $\Tilde{q} = \{q_i\}_{i=1}^{|\Tilde{q}|}$ which is represented as a sequence of words ($q_i$),
and all the candidate answers extracted from the $h$-hop 
augmented KG subgraph surrounding a topic entity, which we denote as $A = \{a_i\}_{i=1}^{|A|}$. Our \PBAMnet model maps them into a joint embedding space, which can capture their latent semantic meanings.
Given the representation (or embedding) of the expanded question $\Tilde{q}$, denoted $\Tilde{\vec{q}}$, and the representations of candidate answers, denoted $\vec{a}_i$ for answer $a_i$, we compute the matching score $S_\text{qa}(\Tilde{q}, a_i)$ between every question and potential answer pair via their dot-product $S_\text{qa}(\Tilde{q}, a_i) = \Tilde{\vec{q}}\bullet \vec{a}_i$. 
The candidate answers are then ranked by their scores.

In the training phase, we force positive candidate answers to
have higher scores than negative candidate answers
by using a triplet-based loss function:
\begin{equation*}
L =
\sum_{\substack{a^+ \in A^+,a^- \in A^-}}
\ell\bigl(S_\text{qa}(\Tilde{q}, a^+), S_\text{qa}(\Tilde{q}, a^-)\bigr)
\end{equation*}
where $\ell(y, \hat{y})=\max(0, 1 + \hat{y} - y)$ is a hinge loss function, and
$A^+$ and $A^-$ denote the positive and negative answer
sets, respectively.
Positive answers are the ground-truth answers, while negative answers are randomly sampled from the candidate answer set.
At testing time, considering there can be multiple answers to a given question, we keep those candidates whose scores are close to the highest score, within
a certain margin $\theta$, as good answers. 
Note that $\theta$ is a hyper-parameter which controls the degree of tolerance.

\subsection{Recommendation with Food Logs}\label{sec:food_rec_with_food_log}

Food recommendation systems often utilize a user's eating history~\cite{forbes2011content,ueda2014recipe} to produce recommendations more aligned to a user's tastes. In addition to personalization through query expansion with profile based likes/dislikes and guidelines, we also support the concept of personalization based on a user food-log by leveraging recipe embeddings~\cite{li2020}.

\subsubsection{Food Log Generation}
Food logs are intended to reflect a user's eating habits. Since we do not have access to extensive food logs that match the recipes in our KG, we create simulated food logs, and utilize them for further personalizing our system's responses.
We focus on five diets to simulate a varied and distinct eating pattern in the food logs: Mediterranean~\cite{migala}, DASH (Dietary Approaches to Stop Hypertension)\cite{bandurski_2018, west}, Vegan~\cite{peta}, Indian~\cite{sadhukhan}, and Keto~\cite{eenfeldt_scher, migala_keto}. To create the simulated food logs, we first construct a list of search terms relevant to recipes and ingredients found in foods associated with each of these diets. Through this approach we can search through all the recipe names in our KG for recipes that highlight these food items or ingredients. We iteratively generate six logs for each diet type by sampling the search results of ten random search terms for that diet. Via this process, we obtain 30 simulated food logs spanning the 5 diet types, with 26 recipes on average.

\subsubsection{Recipe Similarity}\label{sec:recipe_similarity}
Determining the most similar recipes to the user's food log requires a representation of all recipes. We utilize recipe embeddings~\cite{li2020} to obtain a distributional representation of all recipes. Next, for each recipe, we construct a  10-nearest neighbor graph using cosine similarity.
In addition to the food log, we utilize the above 10-nearest neighbor graphs to identify the top $k$ most similar recipes to the recipes in the user's log.
More specifically, we construct this list by taking the $k$ recipes with the most similarity to the food log, where the similarity between a recipe and the food log (denoted as $S_\text{sim}(g, a_i)$) is computed as the largest cosine similarity between the recipe and all the recipes in the food log.

\subsubsection{KG Subgraph Expansion}

In order to consider the aforementioned top $k$ most similar recipes as additional candidate answers for food recommendation, 
we proceed to expand the original KG subgraph (i.e., surrounding the food tag entity) by incorporating those recipes and their associated ingredient and nutrient information.
Notably, we can directly apply \PBAMnet to the expanded KG subgraph without any modification,
and \PBAMnet is able to compute a confidence score $S_\text{qa}(\Tilde{q}, a_i)$ for each pair of expanded question $\Tilde{q}$ and recipe $a_i$.

\subsubsection{Answer Ranking}
When ranking candidate recipes for recommendation, in principle, we prefer those recipes that
not only satisfy various constraints contained in the expanded question (as before),
but also are similar to the user's food log.
To this end, we proceed to compute an overall score $S(\Tilde{q}, f, a_i)$ combining both the KBQA confidence score and the recipe similarity score:
\begin{equation}\label{eq:overall_score}
    \begin{aligned}
S(\Tilde{q}, f, a_i) = (1 - \lambda) \cdot \widetilde{S}_\text{qa}(\Tilde{q}, a_i) + \lambda \cdot \widetilde{S}_\text{sim}(f, a_i)
	 \end{aligned}
\end{equation}
where $\lambda$ is a hyperparameter for controlling the importance of recipe similarity.
$\widetilde{S}_\text{qa}(\Tilde{q}, a_i)=(S_\text{qa}(\Tilde{q}, a_i) - S_\text{min})/(S_\text{max} - S_\text{min})$ is the normalized KBQA score, and $S_\text{qa}(\Tilde{q}, a_i)$ is the unnormalized KBQA score defined in~\cref{sec:kbqa_training}.
$\widetilde{S}_\text{sim}(f, a_i)=(S_\text{sim}(f, a_i) + 1)/2$ is the normalized cosine similarity between food log $f$ and recipe $a_i$, and $S_\text{sim}(f, a_i)$ is the cosine similarity defined in~\cref{sec:recipe_similarity}.
Note that $S_\text{max}$ and $S_\text{min}$ are the maximal and minimal KBQA scores among KBQA scores of all candidate recipes for question $\Tilde{q}$.
During the inference stage,
we use a threshold $\theta$ for selecting good answers.

\section{Experiments}

We evaluate our proposed \PBAMnet model against competitive baseline methods on the personalized food recommendation benchmark.  
The dataset as well as the implementation of our model are publicly available at \url{https://github.com/hugochan/PFoodReq}.

\subsection{Baseline Methods}

Our main baseline is BAMnet~\cite{chen2019bidirectional}, a state-of-the-art IR-based KBQA method
that directly retrieves answers from a KG subgraph by jointly embedding the question and KG subgraph.
We also include a sophisticated embedding-based KBQA method called MatchNN which was motivated by~\cite{bordes2014question}.
To demonstrate that this is a non-trival task, we further include a Bag-of-Word (BOW)~\cite{manning2008introduction} vectors based KBQA baseline.
For both MatchNN and BOW, the candidate answer is encoded as the sum of the corresponding answer type, answer path and answer context embeddings.
When encoding a question (or KG aspect such as answer type), MatchNN employs a bidirectional LSTM~\cite{hochreiter1997long} for encoding the sequence of tokens, 
and BOW simply takes the average of their pretrained word embeddings~\cite{pennington2014glove}.
Note that none of the above baselines explicitly handle personalized QA, and have access to personalized requirements such as dietary preferences and health guidelines. 
They are included as non-personalized baselines in our experiments.

To examine whether our proposed techniques are generic  and can be applied to extend regular KBQA methods for handling personalized food recommendation, we develop personalized versions of the aforementioned three KBQA baselines by leveraging our proposed techniques including query expansion, KG augmentation and constraint modeling.
Therefore, we have three personalized methods that include P-BOW, P-MatchNN and \PBAMnet (i.e., personalized BAMnet)
where \PBAMnet is our main model. 

\subsection{Data and Metrics}

FoodKG~\cite{haussmann2019foodkg} is a large-scale KG that integrates recipes, food ontologies and nutritional data.
It comprises about one million recipes (sourced from~\cite{marin2018learning}), 7.7 thousand nutrient records (sourced from the USDA: \url{www.usda.gov}), and 7.3 thousand types of food (sourced from~\cite{griffiths2016foodon}). 
Overall, it has over 67 million triples. 
In this work, we use FoodKG as the underlying food KG.
ADA Lifestyle Management~\cite{american20195} includes ADA's current clinical practice recommendations and is intended to provide the components of diabetes care, general treatment goals and guidelines, and tools to evaluate quality of care. 
We carefully select food-related guidelines on nutrient and micronutrient needs for ``healthier'' food recommendation.
Following the steps described in~\cref{sec:benchmark_creation}, we create a food recommendation benchmark based on FoodKG and ADA guidelines.

Following previous work on KBQA~\cite{berant2013semantic,chen2019bidirectional}, macro F1 scores (i.e., the average of F1 scores over all questions) are reported
on the test set.
Besides F1, we also report Mean Average Precision (MAP) and Mean Average 
Recall (MAR) scores that are common metrics for evaluating recommender systems \cite{manning2008introduction}.
Note that for all metrics, we consider all the answers returned by a system.


\subsection{Model Hyperparameter Settings}\label{sec:model_settings}

When constructing the vocabularies of words (from questions and KG), entity types (from KG) and relation
types (from KG), we only consider those questions and their corresponding KG subgraphs appearing in the training set.
The vocabulary sizes of words, entity types and relation types are 7009, 7 and 11, respectively.
Given a topic entity, we extract its 2-hop subgraph
(i.e., $h=2$) to collect candidate answers, which is sufficient for our benchmark. 
We initialize word embeddings with 300-dimensional pre-trained GloVe vectors~\cite{pennington2014glove}.
The batch size is set as 32.
The answer module threshold $\theta$ is set to 0.9 and 0.2 for \PBAMnet and \PBAMnetSim, respectively.
The value of $\theta^G$ is set to 0.2 (see \cref{sec:exp_rec_food_log}).
The value of $\lambda$ is set to 0.3.
In the training process, we use the Adam optimizer
\cite{kingma2014adam}. We use $d_m=40$ for constraint modeling.
All hyperparameters are tuned on the development set; for BAMnet we use the default values mentioned in Chen et al.~\shortcite{chen2019bidirectional}.

\begin{table}[ht!]
\vspace{-0.1in}
\centering
\caption{Experimental results (in \%) on the test set.}
\label{table:exp_results}
\vspace{-0.1in}
\begin{tabular}{lllll}
\toprule
 Method  & MAP & MAR & F1\\
  \midrule
 BOW & 2.1 & 2.0 & 2.3\\ 
 MatchNN & 2.7 & 2.7 & 3.0\\
 BAMnet & 3.1 & 3.0 & 3.1\\ 
 \midrule
 P-BOW & 4.5 & 4.4 & 4.2\\
 P-MatchNN & 45.5 & 45.1 & 41.2\\
 \midrule
 \PBAMnet & \bf{62.7} & \bf{61.8} & \bf{63.7}\\ 
 \bottomrule
\end{tabular}
\vspace{-0.2in}
\end{table}

\subsection{Experimental Results}\label{sec:exp_results}

Table~\ref{table:exp_results} shows the evaluation results comparing our method against other baselines.
We observe that the personalized systems significantly outperform their non-personalized counterparts, which verifies the effectiveness of our proposed techniques for 
utilizing the personal information (i.e., dietary preferences and health guidelines) to provide more accurate and personalized food recommendations (regardless of the baseline system used).
Our \PBAMnet model consistently outperforms all models. It overshadows BAMnet by a significant margin (59.6\% absolute improvement in MAP, 58.8\% absolute improvement in MAR, and 60.6\% absolute improvement in F1 ) in all evaluation metrics. It also outperforms P-MatchNN, with an average improvement of 18.8\% absolute improvement over all metrics. 
Moreover, we can observe that there is still scope for improvement on this benchmark, which requires a certain level of reasoning to understand the different constraints in a query and to satisfy them based on the corresponding KG subgraph. While \PBAMnet performs the best, the results indicate that our benchmark is quite challenging and can spur further research.

\subsection{Human Evaluation}


We performed human evaluation for food recommendations by presenting a set of eight evaluators with fifty questions from a randomized test, along with the user personas spanning ingredient preferences (likes and dislikes) and applicable nutrition guidelines. For each question, we provide the answers from BAMnet, P-BOW, P-MatchNN and \PBAMnet models in a randomized order.

\begin{table}[!ht]
\vspace{-0.1in}
\begin{center}
\caption{Human evaluation results ($\pm$ standard deviation) on the test set. The rating scale is from 1 to 10 (higher scores indicate better results).}
\label{table:human_evaluation_results}
\vspace{-0.1in}
\scalebox{1}{
\begin{tabular}{lll}
\toprule
Method &  Score \\
  \midrule
BAMnet & 3.82 (0.70)\\
P-BOW & 4.69 (0.47)\\
P-MatchNN & 7.05 (0.44)\\
\midrule
\PBAMnet & {\bf 7.76} (0.50)\\
\bottomrule
\end{tabular}
}
\end{center}

\vspace{-0.1in}
\end{table}

Each answer presented to the user contains the top three recipes (if more than three recipes were retrieved), the ingredient list, and the nutritional values. The evaluators were instructed to rank the answers in the order of their (subjective) preference on a scale from $1-4$, with $1$ being the best, indicating how well they think the recommended answer set satisfies the various constraints mentioned in the input, without knowing which of the four systems generated the answers.
We adopt a simple rank aggregation strategy to compute the overall score for each system based on the
ranks provided by human evaluators.
Specifically, we map the ranks 1, 2, 3 and 4 to the corresponding scores 10, 7, 4 and 1, so that systems with better ranks have higher scores.
Evaluation scores from all 8 evaluators are collected and averaged as final scores. 
As a result, the highest score a system can get is 10 while the lowest score is 1.
As shown in \cref{table:human_evaluation_results}, \PBAMnet achieves the best result. Furthermore, all of the personalized baselined fared better than the non-personized BAMnet approach. These results conclusively demonstrate the general applicability and effectiveness of our personalization techniques (independently of the method).

\subsection{Model Analysis}

\begin{table}[!t]
\vspace{-0.1in}
\centering
\caption{Ablation study (in \%) on the test set.}
\label{table:ablation_results}
\vspace{-0.1in}
\begin{tabular}{lllll}
\toprule
  Method & MAP & MAR  & F1\\
  \midrule
 \PBAMnet &\textbf{62.7} & \textbf{61.8} & \textbf{63.7}\\ 
 \qquad \textrm{--} KA & 58.5 & 57.8 & 58.6\\ 
 \qquad \textrm{--} CM & 29.7 & 29.3 & 25.9\\ 
 \qquad \textrm{--} QE & 4.5 & 4.5 & 4.2\\ 
 \bottomrule
\end{tabular}
\vspace{-0.1in}
\end{table}

\subsubsection{Ablation Study}

We perform an ablation study to systematically investigate the impact of different model components for our \PBAMnet method.
Here we briefly describe the ablated systems: 
\textrm{--} CM removes the Constraint Modeling module,
\textrm{--} KA removes the KG Augmentation module,
and \textrm{--} QE removes the Query Expansion module.
Note that for the ablated systems, we remove one module each time.
Table~\ref{table:ablation_results} shows the experimental results of the ablated systems as well as the full model.
It confirms our finding that all of the three techniques introduced for handling the personalized recommendation setting greatly contribute to the overall model performance.
Among them, the Query Expansion module contributes most, which demonstrates the  importance of incorporating rich personal information in order to provide accurate recommendation.
By removing the QE module, we observe an average 58.3\% performance drop across all evaluation metrics.
The CM module is also important as we see an average 34.4\% performance drop across metrics if it is removed.
This shows the importance of handling negations for personalized food recommendation.
Lastly, the KA module also contributes to the overall model performance.

\begin{table}[tbh]
\centering
\footnotesize
\vspace{-0.1in}
\caption{Predicted answers by various methods.}
\label{table:case_study}
\vspace{-0.1in}
\begin{tabular}{p{3in}}
\toprule
\multicolumn{1}{c}{\bf Test Query}\\
\midrule
\textbf{Query:} Could you recommend eggplant dishes which do not contain ketchup?\\
\textbf{Preference:} Likes: [``white wine vinegar''], Dislikes: [``sesame'', ``red peppers'']\\
\textbf{Guideline:} Protein: 5.0 g to 30.0 g\\
\textbf{Gold:}  \textbf{[``Moussaka (Vegan or Vegetarian, You Choose.)'']}\\
\toprule
~\\[-0.15in]
\multicolumn{1}{c}{\bf Baseline Methods}\\
\midrule
\textbf{BOW:} [ ``Yunnan Style Grilled Veggies With Dipping Sauce'' ]  \\
\textbf{MatchNN:} [ ``Sicilian Eggplant Parmagiana With Spinach'', ``Savannah Dip'', ``The Real Melitzanosalata'', ``Eggplant, Tomato and Mozzarella Stacks, Hot or Cold'', ``The Ultimate Veggie Burger Thai Style'', ``Rigatoni With Roasted Eggplant Puree'', ... ] \\
\textbf{BAMnet:} [ ``Sicilian Eggplant Parmagiana With Spinach'', ``Savannah Dip'', ``The Real Melitzanosalata'', ``Eggplant, Tomato and Mozzarella Stacks, Hot or Cold'', ``The Ultimate Veggie Burger Thai Style'', ``Rigatoni With Roasted Eggplant Puree'', ... ]  \\
\textbf{P-BOW:} [ ``Black Cherry Tomatoes and Eggplant Pasta Sauce'', ``Italian Sausage and Eggplant Soup'', ``Clarica\'s Melitzanosalata'', ``Caponata'', ``Greek Eggplant Dip'', ... ]\\
\textbf{P-MatchNN:} [ ``White Bean Ratatouille'' ]\\
\toprule
~\\[-0.15in]
\multicolumn{1}{c}{\bf Ablated Methods}\\
\midrule
\textbf{\BlackPBAMnet w/o QE:} [ ``Sicilian Eggplant Parmagiana With Spinach'', ``Savannah Dip'', ``The Real Melitzanosalata'', ``Eggplant, Tomato and Mozzarella Stacks, Hot or Cold'', ``The Ultimate Veggie Burger Thai Style'', ``Rigatoni With Roasted Eggplant Puree'', ... ] \\
\textbf{\BlackPBAMnet w/o KA:} [ ``White Bean Ratatouille'' ]   \\
\textbf{\BlackPBAMnet w/o CM:}
[ ``Moussaka (Vegan or Vegetarian, You Choose.)'', ``White Summer Lasagna'', ``Pasta With Eggplant and Pork'']\\
\toprule
~\\[-0.15in]
\multicolumn{1}{c}{\bf \BlackPBAMnet Method}\\
\midrule
\BlackPBAMnet: [ ``Moussaka (Vegan or Vegetarian, You Choose.)'' ]\\
\bottomrule
\end{tabular}
\end{table}

\subsubsection{Case Study}

Table~\ref{table:case_study} shows the predicted 
answers from \linebreak \PBAMnet and its ablated variants, and other baselines on a testing example.
In this example, the user is asking for eggplant dishes that do not contain ketchup. The user's persona includes both likes (e.g., white wine vinegar) and dislikes (e.g., red peppers), and a health/diet guideline for protein intake.
As we can see, \PBAMnet retrieves the ground-truth response from the KG while other systems make many imprecise recommendations.
For example,  
\PBAMnet w/o CM recommends dishes that contain ``red peppers'' such as ``Pasta With Eggplant and Pork''
when the user dislikes ``red peppers''.
The same mistake is also made by other systems like \PBAMnet w/o QE, MatchNN, and BAMnet that recommend ``The Real Melitzanosalata'' (which also contains red peppers).
Without effectively utilizing the health guideline, many systems also recommend dishes that have protein values outside of the stipulated range.

\begin{table}[thb]
\vspace{-0.1in}
\centering
\caption{Experimental results (in \%) on the test set with and without using recipe similarity.}
\label{table:kbqa_sim_exp_results}
\vspace{-0.1in}
\begin{tabular}{lllll}
\toprule
 Method  & MAP & MAR & F1\\
  \midrule
 \PBAMnet & 27.3 & 26.4 & 32.6 \\ 
 \PBAMnetSim & \textbf{34.5} & \textbf{33.0} & \textbf{36.6} \\ 
 \bottomrule
\end{tabular}

\end{table}

\subsection{Recommendation considering Food Logs}
\label{sec:exp_rec_food_log}

We demonstrate the ability of our food recommendation system in utilizing food logs.
We reuse the pretrained \PBAMnet model evaluated in~\cref{sec:exp_results}.
In order to adapt it to utilize the food log information, we only need to enhance it with the \emph{KG subgraph expansion} and \emph{answer ranking} techniques introduced in~\cref{sec:food_rec_with_food_log} during the inference phase.
We call the resulting system \PBAMnetSim.
For empirical comparison we generate new ground-truth answers that respect both the constraints (as before) and the recipe similarities, using the following strategy.
For each example in the test set,
we combine the ground-truth KBQA answer set $R_\text{qa}$ and the top $k$ similar recipes $R_\text{sim}$ to obtain a potential ground-truth answer set $R_p$.
Then for each recipe in $R_p$, we compute an overall score by combining its KBQA score (i.e., 1 if all constraints are satisfied, and 0.5 if partial constraints are satisfied; otherwise, it is 0) and its recipe similarity score using~\cref{eq:overall_score}.
Finally, we select a subset of recipes (denoted as $R$) from $R_p$ based on their scores using a margin threshold $\theta^G$ following the same strategy mentioned in~\cref{sec:kbqa_training}.


As shown in \cref{table:kbqa_sim_exp_results}, \PBAMnetSim significantly outperforms \PBAMnet across all evaluation metrics. This is because 
\PBAMnetSim additionally considers the similarity between the candidate recipe and the user's food log, and can recommend recipes that are more similar to the user's eating history while still satisfying various constraints on ingredients and nutrients.

\section{Conclusions and Future Work}

In this work, we treated the task of personalized food recommendation as constrained QA over a food KG,
which considers personalized requirements from dietary preferences, health guidelines, and food logs, as additional constraints, utilizing a user's persona.
Novel techniques were proposed to effectively handle numerical comparisons and negations in QA. 
A benchmark for personalized food recommendation was also created for evaluating food recommendation systems.
Experimental results on the benchmark verified the effectiveness of our proposed method.
Future work could consider automatic ways of mining personal preferences and applicable health guidelines from a user's historical behavioral data.

\begin{acks}
This work is supported by IBM Research AI through the AI
Horizons Network (under the RPI-IBM HEALS Project).
\end{acks}

\bibliographystyle{ACM-Reference-Format}
{\footnotesize
\bibliography{document/WSDM21}
}


\end{document}